\documentclass[11pt,a4paper]{article}
\usepackage[nohyperref]{acl2019}
\usepackage[utf8]{inputenc}
\usepackage{times}
\usepackage{latexsym}
\usepackage{graphics}
\usepackage{tikz}
\usepackage{amsmath}
\usepackage{amssymb}
\usepackage{amsthm}
\usepackage{booktabs}
\usepackage{multirow}
\usepackage{url}
\usepackage{adjustbox}
\usepackage{nicefrac}
\usepackage{cleveref}
\crefname{section}{§}{§§}
\Crefname{section}{§}{§§}

\usepackage[normalem]{ulem}

\aclfinalcopy 
\def\aclpaperid{} 

\title{Probing Representations Learned by Multimodal Recurrent and Transformer
Models}

\author{Jindřich Libovický \\
  Charles University\\
  Faculty of Mathematics and Physics \\
  Institute of Formal and Applied Linguistics \\
  {\tt libovicky@ufal.mff.cuni.cz} \\\And
  Pranava Madhyastha \\
  Department of Computing \\
  Imperial College of London \\
  {\tt pranava@imperial.ac.uk} \\}

\date{}

\begin{document}
\maketitle

\begin{abstract}
Recent literature shows that large-scale language modeling provides excellent
    reusable sentence representations with both recurrent and self-attentive
    architectures.
However, there has been less clarity on the commonalities and differences in
    the representational properties induced by the two architectures.  It also
    has been shown that visual information serves as one of the means for
    grounding sentence representations.
In this paper, we present a meta-study assessing the representational quality
    of models where the training signal is obtained from different modalities,
    in particular, language modeling, image features prediction, and both
    textual and multimodal machine translation.
We evaluate textual and visual features of sentence representations obtained
    using predominant approaches on image retrieval and semantic textual
    similarity.
Our experiments reveal that on moderate-sized datasets, \emph{a sentence
    counterpart} in a target language or \emph{visual modality} provides much
    stronger training signal for sentence representation than language
    modeling.
Importantly, we observe that while the Transformer models achieve superior
    machine translation quality, representations from the recurrent neural
    network based models perform significantly better over tasks focused on
    semantic relevance.
\end{abstract}

\section{Introduction}

Conditioning on multimodal information is one of the predominant methods of
grounding representation learned in deep learning models
\citep{chrupala2015learning,lazaridou2015combining}, i.e., relating the word or
sentence representation to non-linguistic real-world entities such as objects
in photographs. In the context of multimodal machine translation (MT), models
using multimodal auxiliary loss have been shown to outperform their text-only
counterparts \citep{elliott2017imagination,helcl2018cuni}. Experiments with
multimodal language models (LMs) also confirm that multimodality influences the
semantic properties of learned representations
\citep{poerner2018interpretable}.

On the other hand, recent experiments with large-scale language modeling
suggest that these models provide sufficiently informative representations
reusable in most natural language processing (NLP) tasks
\citep{peters2018elmo,devlin2018bert}. Current research has also seen an
increasing trend towards investigation on universality of learned
representations where the learned representations are supposed to contain
sufficient inductive biases for a variety of NLP tasks
\citep{conneau2017supervised,howard2018fine}.

Research in evaluating representations has focused on measuring the correlation
between the similarity of learned representations and the semantic similarity
of words \citep{hill2015simlex,gerz2016simverb} and sentences
\citep{agirre2012semeval,agirre2016semeval}. Work on probing representations
include relating learned representations to existing well-trained models by
finding a mutual projection between the learned representations and evaluating
the performance of the projected representations within the trained model
\citep{saphra2018language}, and observing the effect of changes in the
representation by backpropagating the changes to the input
\citep{poerner2018interpretable}.

Universal sentence representations are typically evaluated on its effects on
downstream tasks. \citet{conneau2018senteval} and \citet{wang2018glue} recently
introduced comprehensive sets of such downstream tasks providing a benchmark
for the sentence representation evaluation. The tasks include various sentence
classification tasks, entailment or coreference resolution. However, the
drawback of these methods is that they require generating representations of
millions of sentences which are later used for a rather time-consuming training
of models for the downstream tasks.

In this paper, we investigate representations obtained specifically from
grounded models using the two predominant sequence modeling architectures: a
model based on recurrent neural networks
(RNN;~\citealp{mikolov2010recurrent,bahdanau2014neural}) and a model based on
the self-attentive Transformer architecture \citep{vaswani2017attention}. We
study the learned representations on aspects of grounding, semantics and the
degree to which some of these representations are correlated, irrespective of
modeling choices. Our main observations are: a) models with access to explicit
grounded information learn to ignore image information; b) grounding accounts
for  \emph{better semantic representations} as it  provides a stronger training
signal and is especially pronounced when a model has access to less training
samples; c) while Transformer based models might have better task performance,
we observe that  \emph{RNN based models capture better semantic information}.

\section{Assessing Contextual Representations}\label{sec:method}
%
In this section, we briefly describe the methods used for extracting
representations and for quantifying the representation qualities: Canonical
Correlation Analysis (CCA) for image retrieval evaluation, and cosine distance
for Semantic Textual Similarity evaluation. Finally, we also use Distance
Correlation (DC) for representation similarity evaluation. Whereas the first
two of them are used for evaluation on downstream tasks, the latter one is only
quantifies mutual similarities of the representations.

\paragraph{Canonical Correlation Analysis.}
We take input as the two sets of aligned representations from two different
subspaces, say $\mathbf{T} = \{\mathbf{t_1}, \ldots, \mathbf{t_n}\}$ and
$\mathbf{V} = \{\mathbf{v_1}, \ldots, \mathbf{v_n}\}$, where $\mathbf{t_i}$ and
$\mathbf{v_i}$ are vector representations. CCA \citep{hotelling1936relations}
finds pairs of directions $\mathbf{w_t}, \mathbf{w_v}$, such that the linear
projections of $\mathbf{T}$ and $\mathbf{V}$ onto these directions, i.e., the
canonical representations $\mathbf{w_t}^{\top}\mathbf{T}$ and
$\mathbf{w_v}^{\top}\mathbf{V}$, are maximally correlated. For, further details
on CCA, we refer the reader to~\citet{hardoon2004canonical}.

The most significant property of CCA for our analysis is that CCA is a
\emph{subspace only method} where we obtain naturally occurring correlations
between two spaces. Importantly, we don't \emph{learn} to align, but obtain
alignments that are potentially present between the two subspaces. Further, CCA
is \emph{affine-invariant} due to its reliance on correlation rather than
orthogonality of direction vectors.

We use CCA over mean-pooled sentence representations and image representations
and obtain two highly correlated projections respectively. CCA and its variants
have been used in previous research to obtain cross-modal
representations~\cite{gong2014improving,yan2015deep}.
We evaluate the projected representations on image retrieval task and report
the recall at 10. Note that we do not backpropagate the correlation to the
network and keep the representation fixed because our goal is not training
towards optimal cross-modal representation but only to asses the (already
trained) sentence representation.

\paragraph{Cosine Distance.}
For evaluation on the STS task, we use cosine distance between of vectors
$\mathbf{t}$ and $\mathbf{v}$:
\[ \text{sim}(\mathbf{t, v}) = 1 - (\mathbf{t} \cdot \mathbf{v})\ /\
{\|\mathbf{t}\|  \|\mathbf{v}\|}.  \]
Following the SentEval benchmark \citep{conneau2018senteval}, we report the
Spearman correlation between the distance and human assessments.

The goal of the STS task is to asses how well the representation capture
semantic similarity of sentences as perceived by humans. Similar to the image
retrieval task, we do not fine-tune the representations for the similarity task
and report the Spearman correlation of the cosine distance between the
representations and the ground-truth similarity.

\paragraph{Distance Correlation.}
Distance correlation (DC) is a measure of dependence between any two paired
vectors of arbitrary dimensions~\cite{szekely2007gabor}. Given, two paired
vectors, $\mathbf{t} \in \mathcal{R}^m$ and $\mathbf{v} \in \mathcal{R}^n$ and
suppose that $\phi_1(t), \phi_2(v)$ and $\phi_3(t,v)$ are the individual
characteristic functions and joint characteristic function of the two vectors
respectively. The distance covariance $\text{dcov}^2(\mathbf{t, v})$ between
$\mathbf{t}$ and $\mathbf{v}$ with finite first moments is a non-negative
number given by:
\begin{equation*}
\int_{\mathcal{R}^{m+n}}\|\phi_3(\mathbf{t,v}) -
    \phi_1(\mathbf{t})\phi_2(\mathbf{v})\|_2^2~\psi(\mathbf{t,v})d\textbf{t}~d\textbf{v}
\end{equation*}
where $\psi(\mathbf{t,v}) :=
\{\|\mathbf{t}\|_m^{1+m}\|\mathbf{v}\|_n^{1+n}\}^{-1}$; $m$ and $n$ are the
dimensionalities of $\mathbf{t}$ and $\mathbf{v}$ respectively. The distance
correlation (DC) is then defined as:
\[ \text{dcorr}(\textbf{t, v}) = \frac{\text{dcov}(\textbf{t,
v})}{\sqrt{\text{dcov}(\textbf{t, t})\text{dcov}(\textbf{v, v})}}.  \]
A detailed description of the DC is beyond the scope of this paper, but we
refer the reader to \citet{szekely2007gabor} for a thorough analysis.

Our use of DC is motivated by the result that DC quantifies dependence measure,
especially it equals zero exactly when the two vectors are mutually independent
and are not correlated.
Also, DC measures both \emph{linear} and \emph{non-linear} association between
two vectors.
We use DC to measure the degree of correlation between different
representations. We are especially interested in studying the degree to which
two independently learned representations are correlated.

\section{Experiments}\label{sec:models}

We examine representations for four types of models: a) LMs; b) image
representation prediction models (Imaginet); c) textual MT\@; and d) multimodal
MT models. For each task, we train models based using RNNs and the
Transformer architecture. In addition, we use training datasets of different
sizes.
All models trained with Neural Monkey\footnote{\url{https://github.com/ufal/neuralmonkey}} \citep{helcl2017neural}.

\subsection{Models}

\paragraph{Language Models.} We trained an RNN LM with a single GRU layer
\citep{cho2014gru} of 1000 dimensions end embeddings of 500. The Transformer LM
\citep{vaswani2017attention} has model dimension 512, 6 layers, 8 attention
heads and hidden layer size 4,096.

\paragraph{Imaginet.} The Imaginet models \citep{chrupala2015learning} predict
image representation given a textual description of the image. The
representations is trained only via its grounding in the image representation.

We use a bidirectional RNN encoder with the same hyperparameters as the
aforementioned LM\@.
The Transformer based Imaginet uses the same hyperparameters as the Transformer
based LM\@. The states of the encoder are then mean-pooled and projected with a
hidden layer of 4,096 and ReLU non-linearity to a 2,048-dimensional vector
corresponding to the image representation from the \texttt{ResNet}
\citep{he2016deep}. For a fair comparison, we use the representation before the
final non-linear projection.

For completeness, We also compare the LMs with ELMo \citep{peters2018elmo}, a
representation based on deep RNN LM with character-based embeddings pre-
trained on a large corpus, of 30 million sentences, and BERT
\citep{devlin2018bert}, a Transformer based sentence representation that is
similar to Transformer based LM. We note however that BERT is trained in a
significantly different procedure than regular LMs.

\paragraph{Textual MT models.}
We trained the attentive RNN based seq2seq model \citep{bahdanau2014neural}
with the same hyperparameters as the RNN Imaginet model, and with the
conditional GRU \citep{firat2016cgru} as the decoder. With the Transformer
architecture, we used the same hyperparameters as for the Imaginet models.

Besides the text-only models, we trained Imagination models
\citep{elliott2017imagination} that combine the translation with the Imaginet
models in a multi-task setup. The model is trained to generate a sentence in
target language and predict image representation at the same time.

With multi-task learning, the model takes advantage from large parallel data
without images and monolingual image captioning data at the same time.
Presumably, the model achieves a superior translation quality by being able to
learn a better source sentence representation. At the inference time, the only
requires the textual input.

\paragraph{Multimodal MT models.} For both RNN and Transformer architectures,
we used the same hyperparameters as for the textual models. As in previous
models, we use last convolutional layer of \texttt{ResNet} as image
representation.

In the RNN setup, we experiment with decoder initialization with image
representation \citep{caglayan2017lium,calixto2017incorporating} and with
doubly attentive decoder with three different attention combination strategies
\citep{libovicky2017attention}. First, we concatenate context vectors computed
independently over the image representation and source sentence; second (flat
attention combination), we compute a joint distribution over the image
convolutional maps and the source encoder; third (hierarchical attention
combination), we compute the context vectors independently and combine
them hierarchically using another attention mechanism.

In the Transformer setup, the multimodal models use doubly attentive decoders
\citep{libovicky2018input}. We experiment with four setups: serial, parallel,
flat and hierarchical input combination. The first two are a direct extension
of the Transformer architecture by adding more sublayers in the decoder. The
latter ones are a modification of the attention strategies for in the RNN
setup.

\subsection{Datasets}

\paragraph{Training data.} To evaluate how the representation quality depends
on the amount of the training data, we train our models on different datasets.
The smallest dataset that is used for all types of experiments is
Multi30k~\citep{elliot2016multi} that consists of only 29k training images with
English captions and their translations into German, French, and Czech.

For monolingual experiments (LM and image representation prediction) we further
use English captions from the Flickr30k dataset \citep{plummer2015flickr30k}
that contains 5 captions for each image, in total 145k. The largest monolingual
dataset we work with is a concatenation of Flickr30k and the COCO dataset
\citep{lin2014coco}, with 414k descriptions of 82k images.

For textual MT, where parallel data are needed, we also consider an
unconstrained setup with additional data harvested from parallel and
monolingual corpora \citep{helcl2017wmt,helcl2018cuni} combined with the EU
Bookshop corpus \citep{tiedemann2012opus}, in total of 200M words.

Multimodal MT models are trained on the Multi30k data only.

\paragraph{Evaluation data.}

We fit the CCA on the 29k image-sentence pairs of the training portion of the
Multi30k and evaluate on the 1k pairs from the test set.

For STS, we evaluate the representations on the SemEval 2016 dataset
\citep{agirre2016semeval}. The test set consists of 1,186 sentence pairs
collected from datasets of newspaper headlines, machine translation
post-editing, plagiarism detection, and question-to-question and
answer-to-answer matching on Stack Exchange data. Each sentence pair is
annotated with a similarity value.

\begin{table}[!ht]

\def\dn{$\downarrow$}
\def\up{$\uparrow$}

\begin{center}\scalebox{0.95}{\begin{tabular}{llccc} \toprule
\multicolumn{2}{l}{Language Model}  & ppl.\dn{} & Img.\up{} & STS\up{} \\ \midrule
\multirow{3}{*}{\rotatebox{90}{RNN}}
 & Multi30k                              & 12.10 & 16.6 & .267 \\
 & Flickr30k                             & 11.80 & 22.4 & .340 \\
 & Flickr30k + COCO                      & 11.80 & 23.0 & .378 \\ \midrule
\multirow{3}{*}{\rotatebox{90}{Trans.}}
 & Multi30k                              & 12.42 & \phantom08.9 & .256 \\
 & Flickr30k                             & 11.87 & 17.6 & .283 \\
 & Flickr30k + COCO                      & 11.69 & 21.0 & .303 \\ \toprule
 \multicolumn{2}{l}{ELMo}                & ---   & 28.4 & .631 \\
 \multicolumn{2}{l}{BERT}                & ---   & 22.4 & .624 \\ \toprule

 \multicolumn{2}{l}{Imaginet} & R\kern-1pt\scalebox{.8}{@}\kern-1pt10\up{} & Img.\up{} & STS\up{} \\ \midrule
\multirow{3}{*}{\rotatebox{90}{RNN}}
 & Multi30k                              & 29.5  & 24.4 & .401 \\
 & Flickr30k                             & 37.8  & 26.3 & .483 \\
 & Flickr30k + COCO                      & 39.4  & 25.4 & .501 \\ \midrule
\multirow{3}{*}{\rotatebox{90}{Trans.}}
 & Multi30k                              & 25.5 & 22.1 & .338 \\
 & Flickr30k                             & 36.6 & 29.5 & .436 \\
 & Flickr30k + COCO                      & 38.4 & 28.0 & .451 \\ \toprule

 \multicolumn{2}{l}{Textual MT} & B\scalebox{0.8}{LEU}\up{} & Img.\up{} & STS\up{} \\ \midrule
\multirow{4}{*}{\rotatebox{90}{RNN}}
 & Textual                               & 36.7 & 22.5 & .527 \\
 & Textual U                             & 38.7 & 21.8 & .621 \\ \cmidrule(lr){2-5}
 & Imagination                           & 36.8 & 20.1 & .550 \\
 & Imagination U                         & 38.2 & 27.4 & .622 \\ \midrule
\multirow{4}{*}{\rotatebox{90}{Transformer}}
 & Textual                               & 38.3 & 18.8 & .374 \\
 & Textual U                             & 40.4 & 21.3 & .509 \\ \cmidrule(lr){2-5}
 & Imagination                           & 39.2 & 26.5 & .433 \\
 & Imagination U                         & 42.6 & 31.9 & .512 \\ \toprule

 \multicolumn{2}{l}{Multimodal MT} & B\scalebox{0.8}{LEU}\up{} & Img.\up{} & STS\up{} \\ \midrule
\multirow{4}{*}{\rotatebox{90}{RNN}}
 & Decoder init.\                        & 36.9 & 16.6 & .536 \\
 & Att.\ concatenation                   & 35.7 & 11.4 & .429 \\
 & Flat att.\ comb.\                     & 34.6 & 14.6 & .487 \\
 & Hierar.\ att.\ comb.\                 & 37.6 & 16.7 & .553 \\ \midrule
\multirow{4}{*}{\rotatebox{90}{Transformer}}
 & Serial att.\ comb.\                   & 38.7 & 15.8 & .383 \\
 & Parallel att.\ comb.\                 & 38.6 & 16.8 & .398 \\
 & Flat att.\ comb.\                     & 37.1 & 16.6 & .385 \\
 & Hierar.\ att.\ comb.\                 & 38.5 & 14.3 & .346\\ 
\bottomrule
\end{tabular}}
\end{center}

    \caption{Recall at 10 for image retrieval (`Img.') and Spearman correlation
        for the Sentence similarity task (`STS') for representation extracted
        the models. `U' denotes use of the unconstrained dataset. The first
        column contains task specific metrics on the Multi30k test set: LM
        perplexity, image Recall at 10 and BLUE score,
    resepectively.}\label{tab:quantitative}

    \vspace*{-0.3cm}
\end{table}

\section{Results \& Discussion}\label{sec:results}

\begin{figure}

    \includegraphics{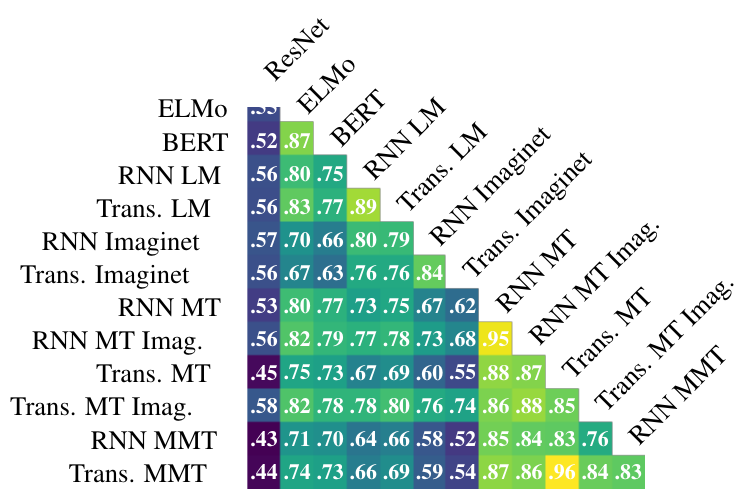}

    \caption{Distance correlation of representations from pairs of selected
    models.}\label{fig:dist_corr}

\end{figure}

We present image retrieval and STS along with the task-specific metrics in
Table~\ref{tab:quantitative}. We observe that on moderately sized datasets,
models conditioned on target language and visual modality provide a stronger
training signal for learning sentence representations than models trained with
simple language modeling objective.

The unconstrained variant of the RNN MMT models obtains a similar performance
in the STS as the ELMo and BERT models even though the training samples was
\emph{orders of magnitude fewer}.

We also observe that while the Transformer based models achieve a superior
translation quality on the MT tasks, the results on STS suggest that RNN models
obtain semantically richer representations. While the textual RNN translation
models perform better on image retrieval than the Transformer models, but the
other way round with Transformer based Imagination models that are explicitly
trained to predict the image representation perform better than their RNN
counterparts. With these consistent observations, we posit that the
Transformer based models, while achieving good performance on the task it is
trained for, seem to ignore image information.

The slight difference between the image retrieval performance of the Imaginet
and Imagination models suggest that training the representation using the
vision and the target language signal is complementary.

We also evaluated the STS performance of the representations with the CCA
projections. The Spearman's correlation is consistently worse by about
$0.02-0.03$.

The encoder of the multimodal MT models that explicitly use the visual input in
the decoder achieve significantly lower image retrieval scores. This
observation suggests that the textual encoder seems to ignore information about
visual aspects of the meaning as the decoder has full access to this
information from the explicit conditioning on image representations. This
observation is in line with the conclusions of the adversarial evaluation
\citep{elliott2018adversarial,libovicky2018input}.

\begin{figure}

    \includegraphics{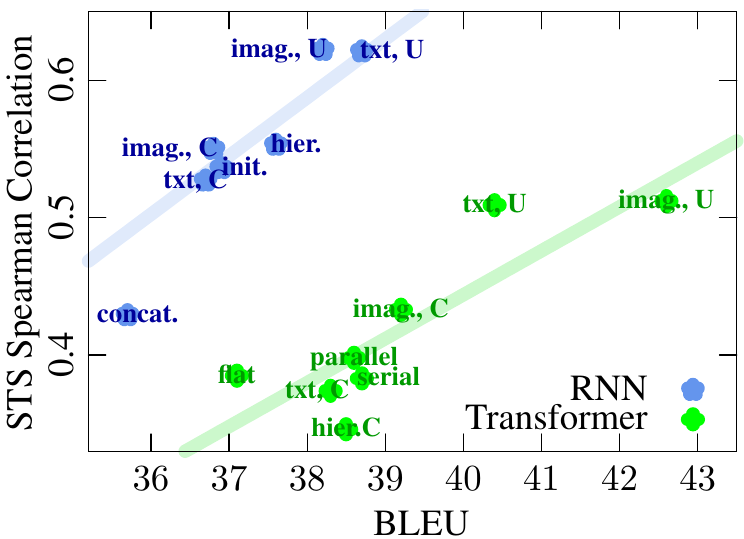}

    \caption{Plot of dependence of the BLEU score on the Spearman correlation
    on the STS dataset.}\label{fig:bleu_vs_sts}

\end{figure}

Our experiments also indicate that the performance on STS is highly correlated
with the translation quality for both the RNN based and the Transformer based
models (see Figure~\ref{fig:bleu_vs_sts}) which is in contrast in findings
of \citet{cifka2018bleu} who measured correlation of BLEU score and STS under similar conditions. In addition, we observe that
Transformers  perform significantly worse with STS than their RNN counterparts.
The translation quality also appears to be highly correlated with the amount of
available training data and image retrieval abilities of the representation
(see Table~\ref{tab:correl}).

\begin{table}
\begin{center}\scalebox{1.0}{\begin{tabular}{lcc}
\toprule
Correlation of BLEU and \ldots & Trans. & RNN \\ \midrule
Image retrieval R@10    & .825 & .700 \\
STS performance         & .852 & .873 \\
Training data size      & .867 & .724 \\ \bottomrule
\end{tabular}}\end{center}

\caption{Pearson correlation of MMT performance and representation
    properties.}\label{tab:correl}

    \vspace*{-0.2cm}
\end{table}

The result of DC for selected models are shown in Figure~\ref{fig:dist_corr}.
The DC of the image and the sentence representations is proportional to the
image retrieval score, also, images have the least correlation distance
resulting in poorer resultant CCA based projections. Sentence representations
seem to be more similar among the tasks than among the architectures. Most
notable is the mutual similarity of representation from all MT systems
regardless of the architecture and the modality setup.

\section{Conclusions}

We conducted a set of controlled and thorough experiments to asses the
representational qualities of monomodal and multimodal sequential models with
predominant architectures. Our experiments show that grounding, in either the
visual modality or with another language, especially their combination in the
Imagination models, results in better representations than LMs trained on
datasets of similar sizes. We also showed that the translation quality of the
MT models is highly correlated both, with the ability of the models to retain
image information and with the semantic properties of the representations.

The RNN models tend to perform better on both the semantic similarity and image
retrieval tasks, although they do not reach the same translation quality. We
hypothesize this is because of the differences in the architecture that allows
the Transformer network to directly access information that the RNN needs to
pass in its hidden states.

\section*{Acknowledgement}

Jindřich received funding from the Czech Science Foundation, grant no. 18-02196S.

\bibliography{references}
\bibliographystyle{acl_natbib}

\end{document}